\documentclass[11pt,a4paper]{article}

\usepackage[utf8]{inputenc}
\usepackage[T1]{fontenc}
\usepackage{mathptmx}
\usepackage[margin=2.6cm]{geometry}
\usepackage{amsmath,amssymb}
\usepackage{booktabs}
\usepackage{array}
\usepackage{microtype}
\usepackage[numbers,super,sort&compress]{natbib}
\usepackage[colorlinks=true,linkcolor=blue!50!black,citecolor=blue!50!black,urlcolor=blue!50!black]{hyperref}
\usepackage{setspace}
\usepackage{xcolor}
\usepackage{titlesec}

\titleformat{\section}{\large\bfseries}{\thesection.}{0.6em}{}
\titleformat{\subsection}{\normalsize\bfseries}{\thesubsection}{0.6em}{}

\newcommand{\Itar}{I_{\mathrm{target}}}
\newcommand{\Ihum}{I_{\mathrm{human}}}
\newcommand{\Iind}{I_{\mathrm{indiv}}}
\newcommand{\lolli}{\multimap}
\newcommand{\oc}{{!\,}}
\newcommand{\Ctx}{\mathrm{Ctx}}
\newcommand{\Obs}{\mathrm{Obs}}
\newcommand{\Rep}{\mathrm{Report}}

\title{\bfseries Identity from the Outside:\\ A Conceptual Framework and Research Program\\ for AI Personality Clones}
\author{Luc E. Brunet\\[2pt] \normalsize R\&D Mediation}
\date{\normalsize Preprint --- v2, July 2026}

\begin{document}

\maketitle

\begin{abstract}
\noindent AI ``personality clones'' force a re-examination of personal identity in operational terms. Setting aside the hard problem of consciousness as currently intractable, we adopt a deliberately operationalist stance: identity is approached through the indiscernibility of manifestations, as assessed by an observer, over a duration. We first distinguish three criteria that the word ``identity'' conflates --- fidelity to a target person ($\Itar$), generic human-likeness ($\Ihum$), and individuality ($\Iind$) --- and argue that existing evidence bears very unevenly on them. We then propose a six-term factorization of observed identity --- generative substrate, dispositions, memory, update dynamics, context, and exogenous contingencies --- presented as a heuristic first factorization with acknowledged overlaps, together with a state-space formulation. Indiscernibility is defined as one minus a judge's distinguishing advantage under an explicit protocol, and the ``coefficients'' of the factorization are reinterpreted as local sensitivities, with interactions, estimable by randomized ablation. The central claim is downgraded from a theorem to a conditional conjecture: under explicit hypotheses about the agent's information regarding its own persistence and about consequences bearing on its own stakes, versionability tends to degrade long-horizon indiscernibility. A formal analogy with $\lambda$-calculus, linear typing, and bisimulation is developed as a disciplined analogy --- clarifying what linearity does and does not establish --- and a thermodynamic section is recast as a resource-accounting analogy. Between the product-clone and the individual we identify a third object, the \emph{delegate}: a task-limited, bounded-lifespan partial clone terminating in a bandwidth-limited testament. We review the empirical literature as indirect evidence mapped onto the three criteria, propose an experimental program with explicit causal, statistical, and ethical design, and argue that the correct long-horizon fidelity criterion for a clone is not trajectory fidelity but \emph{climate fidelity}: matching the conditional distribution of a person's possible responses. The best possible clone is the one that diverges from the original as the original would have diverged from itself.
\end{abstract}

\section{Introduction}

\subsection{Routing around the hard problem}

The hard problem of consciousness --- why and how physical processes give rise to subjective experience\cite{chalmers1995} --- remains without an agreed-upon solution, and nothing in the present state of philosophy or neuroscience suggests that AI personality clones will wait for one. We therefore make a deliberately modest move: we do not attempt to solve the hard problem; we route around it. Since consciousness is only ever known through its manifestations, we adopt a stance in the lineage of Turing's criterion\cite{turing1950}: we refrain from ruling on the interiority of a system and approach identity through the \textbf{indiscernibility of manifestations}, whether of biological or artificial origin.

A terminological clarification is in order. What we practice is closer to a \textbf{methodological, relational operationalism} than to externalism in the classical philosophy-of-mind sense, or to functionalism, which also concerns internal causal organization. We use ``from the outside'' as shorthand for: identity attributions are here operationalized through observation protocols, without commitment to what identity ultimately \emph{is}. This stance does not deny that internal structure exists or matters; it commits us to a discipline --- whenever we invoke a hidden difference between systems (as we will, in \S\ref{sec:conjecture} and \S\ref{sec:thermo}), we owe an explicit causal hypothesis connecting that hidden difference to observable manifestations.

This modesty is what makes the problem tractable \emph{in practice}. While fully and indefinitely indiscernible clones may be unattainable (we conjecture below that, under stated conditions, they are), \textbf{plausible partial clones are already accessible today} --- in a precise and limited sense: systems that pass short, generic human-likeness tests under specific protocols, and interview-based agents that reproduce individual survey responses at a substantial fraction of people's own test--retest consistency (\S\ref{sec:empirical}). The interesting question is therefore no longer \emph{whether} something like personality can be cloned, but \emph{which components} of identity are cheap to clone, which resist, why they resist, and what kind of object a deliberately partial clone is.

\subsection{Three criteria, not one}
\label{sec:criteria}

The word ``identity'' conflates at least three questions that this paper --- and the literature --- must keep separate:

\begin{enumerate}
\item \textbf{$\Itar$ --- fidelity to a target person.} Does this system behave like \emph{this particular person} rather than like someone else? This criterion presupposes a reference (the original, or data about them) and, in its strongest form, a judge who knows the target.
\item \textbf{$\Ihum$ --- generic human-likeness.} Can this system be distinguished from \emph{some} human? This is what classical and modern Turing tests measure\cite{jones2024naacl,jones2024turing,jones2025turing,rathi2024}; no particular person is at stake, and a successful \emph{fictional} persona passes it.
\item \textbf{$\Iind$ --- individuality.} Does this system manifest a trajectory of its own --- stakes, vulnerability, non-cancellable consequences? This criterion, unlike the first, requires no reference: it asks not ``is this the same person?'' but ``is this an individual at all?''.
\end{enumerate}

These criteria are logically independent: a system can be strongly human-like while resembling no one in particular (persona-driven chatbots); it can match a target's questionnaire responses\cite{park2024} without sustaining individuality over time; and one can imagine an unmistakable individual that resembles neither its target nor a typical human. The empirical literature reviewed in \S\ref{sec:empirical} bears almost entirely on $\Ihum$ at short durations and on a questionnaire-based facet of $\Itar$; findings on perceived identity and moral traits\cite{strohminger2014,strohminger2015} concern third-party \emph{judgments} of identity, which is yet another measure. The conjecture at the heart of this paper (\S\ref{sec:conjecture}) concerns primarily $\Iind$ and the intimate-judge, long-horizon form of $\Itar$.

Throughout, we index indiscernibility by the observation protocol: calendar duration, number and depth of interactions, and judge class. Identity, on this approach, admits of degrees\cite{parfit1971}, and the degree is a property of the (system, observer, protocol) triple --- not of the system alone. This also departs from the classical Lockean identification of personal identity with memory\cite{locke1694}, for reasons that turn out to have empirical support (\S\ref{sec:strohminger}).

\subsection{Roadmap}

Section~\ref{sec:factorization} introduces a six-term factorization of observed identity, a state-space formulation, and a conditional impracticability conjecture. Section~\ref{sec:advantage} defines the indiscernibility score as a distinguishing advantage and reinterprets the factorization's coefficients as local sensitivities with interactions. Section~\ref{sec:formal} develops a formal analogy --- $\lambda$-calculus, linear typing, bisimulation --- stated explicitly \emph{as} an analogy, and recasts the thermodynamic argument as resource accounting; it also introduces the \emph{delegate}. Section~\ref{sec:empirical} reviews the empirical literature as indirect evidence mapped onto the three criteria. Section~\ref{sec:program} details an experimental program. Section~\ref{sec:climate} develops climate fidelity as the correct long-horizon criterion. Section~\ref{sec:limits} states limitations and ethical considerations.

\section{A Six-Term Factorization}
\label{sec:factorization}

Starting from a first formulation (identity = language model + personality + memory), progressive refinement leads to six terms:
\begin{equation}
E \;\approx\; S + D + M + U + C + X
\label{eq:additive}
\end{equation}

\begin{table}[ht]
\centering
\small
\begin{tabular}{@{}l l p{7.6cm} l@{}}
\toprule
\textbf{Term} & \textbf{Name} & \textbf{Content} & \textbf{Nature} \\
\midrule
$S$ & Generative substrate & The generative machinery itself --- read broadly: not only a language model but voice, perception, embodiment, action policy & State \\
$D$ & Dispositions & Values, style, traits, characteristic manner of expression & State \\
$M$ & Memory & Culture, biography, recollections, situated knowledge & State \\
$U$ & Update dynamics & How the system is modified by what it lives through: integration of experience, evolution of beliefs, invariants under change & Process \\
$C$ & Context & Informational anchoring in the situation: place, moment, interlocutor, multimodal stream & Input \\
$X$ & Exogenous contingencies & Events with \textbf{persistent, non-cancellable consequences bearing on the system's own goals, resources, or commitments} & Input \\
\bottomrule
\end{tabular}
\caption{The six-term factorization of observed identity. (We write $X$ rather than an earlier $C'$ to avoid confusion with $C$.)}
\label{tab:terms}
\end{table}

\subsection{Status of the factorization}
\label{sec:status}

\textbf{This is a heuristic first factorization, not yet an identifiable model.} Its boundaries are porous, and we prefer to state the overlaps rather than paper over them: the ``manner of learning'', initially placed in $D$, belongs equally to $U$; $M$ evolves under the action of $U$ --- memory is the sediment of the update dynamics; an exogenous event $X$ becomes part of $M$ once integrated; $C$ and $X$ are \emph{inputs}, not processes in the same sense as $U$; affect, goals, bodily constraints, and social commitments are not cleanly localized in any single term --- they plausibly live in a resource/commitment register that interacts with several.

A state-space formulation makes the roles explicit. Writing $\sigma_t$ for the internal state, $c_t$ for context, $x_t$ for exogenous events, and $o_t$ for manifestations:
\begin{equation}
o_t \sim P_S(\,\cdot \mid \sigma_t, c_t\,), \qquad
\sigma_{t+1} \sim K_U(\,\cdot \mid \sigma_t, c_t, x_t\,)
\label{eq:statespace}
\end{equation}
with $\sigma_0 = (D_0, M_0, R_0)$, where $R_0$ is an explicit register of resources, goals, and commitments. Here $S$ is the manifestation policy, $U$ the transition kernel, and the six terms become: two kernels ($P_S$, $K_U$), an initial condition ($D_0, M_0, R_0$), and two input streams ($c_t$, $x_t$). The factorization is \emph{useful} to the extent that these components can be separately intervened on --- which is precisely what the ablation program of \S\ref{sec:program} is designed to test.

\textbf{Two conventions.} First, the ``$+$'' in Eq.~\eqref{eq:additive} is a linear fiction: the terms compose rather than add --- contingencies have identity-shaping effects only \emph{through} $U$; context is read only through dispositions. Second, the state/process line is also, to a first approximation, the easily-clonable/hardly-clonable line: a state can be copied; a trajectory in progress cannot be copied without being forked.

\textbf{On contingency and stakes.} ``Non-cancellable randomness'' is too narrow a gloss for $X$. A biographical contingency can be deterministic yet unforeseeable by the system; conversely, injected stochastic noise is cheap to simulate and identity-inert. What matters is not randomness but \textbf{consequence}: an event counts as $X$ to the degree that it durably alters the system's own resources, goals, or commitments, with no rollback available \emph{to the system}. Contingency alone does not create a stake; it must interact with something the system has to lose.

\subsection{From theorem to conjecture}

An earlier version of this framework stated an ``impracticability theorem'': the more practical a clone (copyable, versionable, resettable), the less indiscernible over time. We now state this more carefully, because as a theorem it is not proven --- and, worse, in its naive form it conflicts with our own operationalist criterion. A dormant checkpoint that is never used, and that the agent knows nothing about, changes no manifestation; under our criterion it cannot, by itself, make a difference. If versionability matters, it must matter \emph{through a causal path to behaviour}.

\subsection{The impracticability conjecture (conditional form)}
\label{sec:conjecture}

We therefore state the claim as a conditional conjecture, with its causal bridge made explicit.

\paragraph{Hypotheses.}
\begin{itemize}
\item[\textbf{(H1)}] \textbf{Self-model coupling.} The agent has information (beliefs, however implemented) about its own persistence regime --- whether its states can be saved, reverted, duplicated --- and this information modulates its behaviour.
\item[\textbf{(H2)}] \textbf{Genuine stakes.} Events in the agent's stream carry persistent consequences for a register of goals/resources/commitments that the agent's update dynamics actually consult.
\item[\textbf{(H3)}] \textbf{Probing judges.} Judges interact over long horizons and can actively probe for the behavioural signatures of stakes: caution, commitment, wear, the weight of past words.
\item[\textbf{(H4)}] \textbf{Costly simulation.} The behavioural signatures in (H3) cannot be cheaply and stably simulated without implementing (H1)--(H2).
\end{itemize}

\paragraph{Conjecture.} Under (H1)--(H4), versionability and resettability tend to degrade long-horizon indiscernibility --- primarily $\Iind$, and $\Itar$ before intimate judges --- because they sever the causal path from consequence to manifestation: an agent whose consequences are revocable, and which can act accordingly, will not durably exhibit the behavioural profile of an agent whose consequences are not.

Three remarks. First, this is much weaker than an in-principle impossibility --- deliberately so; it is a testable causal claim, and \S\ref{sec:program} (experiment 3) is designed to test it, including the crucial dissociation between the agent's \emph{belief} about resettability and the operator's \emph{actual capability}. Second, the conjecture predicts nothing about short horizons, where (H3) has no room to operate --- consistent with the empirical record (\S\ref{sec:turing}). Third, if (H4) fails --- if stake-signatures can be cheaply faked at arbitrary depth --- the conjecture fails, and that would itself be a major finding.

\section{Indiscernibility as Distinguishing Advantage}
\label{sec:advantage}

\subsection{Definition}
\label{sec:definition}

Talk of an ``indiscernibility score'' is empty without a protocol. Let a judge $j$ interact, under a protocol $\pi$ specifying (i) whether the judge sees one system or a matched pair, (ii) whether the judge knows the target person, (iii) an interrogation budget, and (iv) a decision task. Over transcripts generated by the original $P$ and the clone $Q$, define the judge's \textbf{distinguishing advantage} $\operatorname{Adv}_{\pi,j}(P,Q) \in [0,1]$ --- for a binary forced choice, $\operatorname{Adv} = |2 \cdot \text{accuracy} - 1|$, so that a systematically \emph{inverted} judge counts as discriminating, not as fooled; for richer tasks, signal-detection ($d'$, AUC) or proper-scoring analogues apply. Then:
\begin{equation}
I_{\pi,j} \;=\; 1 - \operatorname{Adv}_{\pi,j}(P, Q)
\label{eq:advantage}
\end{equation}
For a class $J$ of judges, one may report the empirical mean (typical-judge indiscernibility) or the infimum over $J$ (adversarial indiscernibility); the two answer different questions, and the gap between them is itself informative. The idealized limit --- indiscernibility against \emph{every} context --- corresponds to the contextual equivalence of \S\ref{sec:obseq}; actual experiments live far from that limit, with bounded, human judges.

\paragraph{Disentangling ``duration''.} Calendar time $\tau$, number of exchanges $k$, interrogation budget, and context diversity are distinct resources and must be separately controlled: ten exchanges spread over six months and ten thousand exchanges in a week are different tests, plausibly probing different modules (calendar time exposes the absence of exogenous change; exchange count exposes inconsistency).

\subsection{Sensitivities, not coefficients}
\label{sec:sensitivities}

Literal coefficients ($\alpha S + \beta D + \dots$) have no dimensional meaning, and the modules interact: ablating memory changes how dispositions are expressed; freezing the update dynamics makes memory incoherent over time. We therefore interpret the quantities of interest as \textbf{local sensitivities}: the drop in $I_{\pi,j}$ when one module is degraded, \emph{at a reference configuration and along a specified quality scale}, i.e.\ $\partial I / \partial(\text{module quality})$ evaluated locally. Estimating them honestly requires randomized factorial ablations with explicit interaction terms --- variance-based indices (Sobol') or cooperative-game attributions (Shapley values) are the appropriate machinery\cite{sobol2001} --- rather than one-at-a-time ablation alone.

Conjectured structure of the sensitivity surface:
\begin{enumerate}
\item \textbf{Dependence on protocol.} At short interaction depth, $S$ and $D$ dominate; at medium depth, $M$ and $C$; at long calendar horizons, $U$ and $X$ --- the trajectory betrays the clone. (Consistent with, though not established by, the evidence in \S\ref{sec:empirical}.)
\item \textbf{Dependence on judge class.} A stranger, a colleague, and a spouse are different instruments; the intimate knows the trajectory, not merely the state.
\item \textbf{An economic frontier.} Short protocols, stranger judges $\rightarrow$ $S + D$ suffice (cheap, current state of the art). Medium $\rightarrow$ $M + C$ (the present engineering front: persistent memory, multimodality). Long horizons, intimate judges $\rightarrow$ $U$ and $X$, which is where the conjecture of \S\ref{sec:conjecture} bites.
\end{enumerate}

A philosophical reading, offered as such: what is easy to clone ($S$, $D$) is what a person shares with their type; what resists ($U$, $X$) is what belongs only to their trajectory. The sensitivities measure the degree of singularity of each component.

\section{A Formal Analogy: $\lambda$-Calculus, Linearity, Bisimulation}
\label{sec:formal}

\paragraph{Status of this section.} What follows is a \emph{disciplined analogy}, not a theory. We map the framework onto established formalisms in order to sharpen its joints --- and we flag, at each step, where the mapping is exact, where it is a modelling choice, and where it breaks. The value of the exercise is not proof but articulation: the formalisms force distinctions (state vs.\ stream, duplicable vs.\ consumed, equivalence vs.\ metric) that ordinary language blurs.

\subsection{Indiscernibility as observational equivalence}
\label{sec:obseq}

The operationalist criterion of \S\ref{sec:criteria} has a formal counterpart: \textbf{observational (contextual) equivalence} in the $\lambda$-calculus\cite{church1936,barendregt1984,morris1968}. Two terms are equivalent iff no context $C[\cdot]$ distinguishes them by observable behaviour. The judge becomes, formally, a \emph{context}, and interaction depth finds its natural place: equivalence bounded to $k$ exchanges defines a hierarchy $\cong_1 \,\supseteq\, \cong_2 \,\supseteq\, \dots \,\supseteq\, \cong_\infty$, and the sensitivity surface of \S\ref{sec:sensitivities} describes which modules cease to be interchangeable at each level. The coinductive variant, \textbf{bisimulation}\cite{milner1999}, is better suited still, being designed for interactive systems that do not terminate.

Two caveats keep the analogy honest. Contextual equivalence and bisimulation coincide only under full-abstraction conditions, which must be checked per calculus, not assumed. And both are Boolean, whereas our subject is graded and stochastic: the appropriate tools are probabilistic transition systems equipped with \textbf{behavioural pseudometrics}\cite{desharnais2004} or depth-bounded distinguishing games. Developing the framework in that quantitative setting is future work; here we use the Boolean notions as the idealized skeleton.

\subsection{Structure replaces summation}

Under the analogy, the six terms rewrite as follows: $S$ is the evaluator; $D$ and $M$ form the environment of a closure --- captured at first application, and thereafter accessible chiefly \emph{through the term's behaviour} rather than as free-standing data (the degree to which implementations expose captured environments varies; the point is about interface, not metaphysics); $U$ is a fold over a stream of events, the full identity being the fixed point of a process; $C$ and $X$ are the input stream. The ``$+$'' thus resolves into: a product type for the initial state, a composition for the reading of context, a coinductive fold for the trajectory. The state/process distinction of \S\ref{sec:status} becomes the induction/coinduction distinction: an interactive identity is defined by what it will answer, indefinitely, to any solicitation. An identity is not a datum; it is a conduct.

\subsection{What the pure $\lambda$-calculus leaves out}
\label{sec:pure}

Three features make the pure $\lambda$-calculus the native formalism of the \emph{practical} clone --- and their absence is exactly what the practical clone lacks:

\begin{enumerate}
\item \textbf{Duplication is free.} Contraction is unrestricted: $\lambda x.\, f\, x\, x$ copies any term at no cost. Every term is clonable.
\item \textbf{Nothing is at stake.} Evaluation is consequence-free: terms can be discarded (weakening), and any saved term can be re-run at will. Checkpoint-and-replay is external to the calculus, costless, and invisible to it. (Note: $\beta$-reduction itself is \emph{not} reversible --- it can erase information; the point is not reversibility of reduction but the absence of any cost, for the term, attached to running, re-running, or discarding it.)
\item \textbf{No exogenous events.} The pure calculus has no primitive for contingency; confluence (Church--Rosser) guarantees that the order of reduction cannot affect the result. Contingency is not ``eliminated'' --- it simply has no way in.
\end{enumerate}

The corrections are well studied. \textbf{Linear logic}\cite{girard1987} restricts contraction and weakening: a hypothesis declared linear is consumed exactly once. \textbf{Effectful calculi} (monads\cite{moggi1991}) make effects explicit in the types; whether a given effect is irreversible depends on the chosen semantics, which is precisely where modelling decisions become visible. \textbf{Coalgebra}\cite{rutten2000,jacobs2016} unifies streams, observation, and bisimulation in one framework and is the natural mathematical home of this program.

\subsection{What linearity does --- and does not --- establish}
\label{sec:linearity}

Here the analogy must be handled with care, because it is tempting to overread.

\textbf{What linearity is.} Linear logic does not \emph{prove} that trajectories are non-clonable. It provides a discipline in which one may \emph{declare} a resource linear and then soundly track the consequences of that declaration. Our modelling claim --- external to the logic --- is that lived events are of this type: the stream of consequential events, as defined by (H2), is consumed exactly once by the system that lives it. Given that declaration, the type system enforces what the declaration means: \textbf{a consumed stream cannot be replayed.}

\textbf{What linearity does not say.} It does not say that the \emph{process} cannot be multiplied. A duplicable program can open many independent linear sessions --- this is what every software service does: copyable code, one linear stream per session. Such a service is well typed and is nonetheless a product, not an individual. The typing therefore locates the crucial difference more precisely than the slogan ``cloning a trajectory is a type error'' (which we retract in that general form): what separates the individual from the service is not stream-linearity, which both have, but whether the \textbf{state} carries non-duplicable stakes --- whether there is a register $R$ (H2) that the events irreversibly move, and that no fresh session restarts. Linearity supplies the grammar of consumption; the stakes hypothesis supplies the content.

\subsection{A typing sketch}
\label{sec:sketch}

With those caveats, the framework can be sketched as follows (sketch, not a verified development; a full treatment requires session types, an affine or explicitly-destructive discipline for termination, and probabilistic semantics):
\begin{align*}
\mathrm{step} &:\ \oc\bigl(\, \Sigma \otimes (\Ctx \otimes X) \lolli \Obs \otimes \Sigma \,\bigr)
  && \text{--- code: duplicable}\\
\sigma_t &:\ \Sigma
  && \text{--- current state: linear}\\
\mathrm{Life} &\;\triangleq\; \nu Y.\ (\Ctx \otimes X) \otimes Y
  && \text{--- open-ended stream (individual)}\\
\mathrm{Task} &\;\triangleq\; \text{inductive, finite stream}
  && \text{--- bounded stream (delegate, \S\ref{sec:delegate})}\\[4pt]
\mathrm{identity} &:\ \Sigma \lolli \mathrm{Life} \lolli (\nu Z.\ \Obs \otimes Z) \\
\mathrm{summarize} &:\ \Sigma \lolli \oc\Rep
  && \text{--- explicit, lossy serialization}\\
\mathrm{merge} &:\ \Sigma \otimes \Sigma \lolli \Sigma
  && \text{--- typable: fusion is possible}
\end{align*}

Key readings. The \emph{code} ($\mathrm{step}$, the pair $S$, $U$) lives under ``$\oc$'': copying the machinery is unrestricted --- this is the software-service fact. The \emph{current state} is linear: there is no $\mathrm{dup} : \Sigma \lolli \Sigma \otimes \Sigma$ in the discipline. An operator who checkpoints a running system is, in this picture, \textbf{adding a duplication rule that the object discipline lacks} --- stepping outside the system's own resource logic. Whether that intervention makes an observable difference is exactly what the conjecture of \S\ref{sec:conjecture} (via H1--H2) claims, and what experiment 3 of \S\ref{sec:program} tests; the type system marks the site of the intervention, it does not adjudicate its effects.

Note also what is \emph{not} forbidden: $\mathrm{merge}$ is a perfectly typable linear function consuming two states and producing one. Reintegration of a fork is therefore not ill-typed; it is lossy and possibly conflictual, but the linear discipline permits it. Similarly, $\mathrm{summarize}$ is required as an \emph{explicit} operation: nothing in the types makes a state's content automatically duplicable --- serializing a trajectory's residue into a shareable report is an act, with a signature, and (as \S\ref{sec:delegate} argues) a bound.

An identity criterion then takes the bounded form: $E_1 \cong_k E_2$ iff no context of interaction depth $\le k$ distinguishes them --- with the hierarchy of \S\ref{sec:obseq}, and with the understanding that the quantitative version (\S\ref{sec:obseq}, caveats) is the one experiments can estimate.

\subsection{Thermodynamics as resource accounting}
\label{sec:thermo}

An earlier version of this section claimed that linear logic's no-duplication, the quantum no-cloning theorem, and the second law were ``three faces of the same constraint''. That claim was too strong, and we withdraw it. The corrected picture is more modest and, we think, more useful.

\textbf{What physics actually provides.} Landauer's principle\cite{landauer1961} assigns a minimal thermodynamic cost to \emph{logically irreversible erasure} of information --- not to copying, and not to experience in general; classical copying can in principle be performed in a logically reversible manner given suitably prepared target memory, the costs being displaced into preparation and reuse of resources. The quantum no-cloning theorem\cite{wootters1982} forbids perfect copying of \emph{unknown quantum states}; it does not apply to a classical digital representation of a personality. Between these results and linear logic there are genuine conceptual and categorical kinships --- each formalizes, in its own domain, a restriction on free duplication or free erasure --- but no general equivalence, and we claim none.

\textbf{The accounting analogy.} What thermodynamics does license is a modest observation: every physical implementation --- product or organism --- involves resources, system boundaries, and irreversibilities; living systems persist by continuously exporting entropy\cite{schrodinger1944}. Dissipation therefore cannot \emph{distinguish} individual from product: everything dissipates. What can distinguish them is \textbf{bookkeeping}: on whose account do the consequences land? A versionable clone is a system whose failures can be silently absorbed by an operator --- reverted, patched, reset --- so that, if (H1)--(H2) hold, its own goal-register never bears the full weight of events. An individual is a system whose register does. On this reading, $X$ is not entropy; $X$ is \textbf{exposure to persistent consequences that the system cannot pass to a third party} --- an accounting property, physically implemented, but not reducible to a dissipation figure. The product/individual frontier, drawn economically in \S\ref{sec:conjecture} and typed in \S\ref{sec:sketch}, is here rephrased in accounting terms; the three framings are consistent, but they are one \emph{hypothesis} viewed from three sides, not three independent proofs.

\textbf{Depth, not incompressibility.} An earlier draft characterized a singular biography as ``incompressible without being random''. As stated, this was wrong: random strings are precisely the ones that maximize algorithmic incompressibility\cite{kolmogorov1965}. The notion needed is Bennett's \textbf{logical depth}\cite{bennett1988} (or related measures of sophistication): a deep object is one whose most plausible generation requires a long computation --- a long history. A biography is \emph{deep}: cheap neither to generate nor to regenerate from a short description, yet far from random, because it has coherence and motifs. Noise is incompressible and shallow; a life is (moderately) compressible and deep. What the delegate's testament loses (\S\ref{sec:delegate}) is depth, not entropy.

\subsection{The delegate: a bounded third object}
\label{sec:delegate}

Between the product (indefinitely versionable, operator-accounted) and the individual (unique, exposed, self-accounted), the framework admits a third object, long anticipated by science fiction under the name of \emph{partials}: the \textbf{delegate} --- a clone deliberately restricted to a bounded task and a bounded lifespan, terminating not by reintegration but by the emission of a final report, a \emph{testament}.\footnote{Science fiction explored this design space narratively and speculatively before any formalization. Greg Bear's \emph{Eon} cycle\cite{bear1985} introduces \emph{partials} --- partial personality copies detached for a task and later reabsorbed by the primary personality; in the terms above, Bear's reintegration is a high-bandwidth $\mathrm{merge}$, possible but conflict-prone, and Bear indeed dramatizes the pathological cases. David Brin's \emph{Kiln People}\cite{brin2002} is the low-bandwidth case exactly: clay duplicates with a fixed $\sim$24-hour lifespan and an optional end-of-life memory upload --- a testament under an explicit bound. Charles Stross's \emph{Accelerando}\cite{stross2005} explores forked selves whose divergence is never healed. The fiction converges on the trade-off the types exhibit: bounded delegation with bounded transmission is workable; unbounded duplication of an open-ended life is where the trouble concentrates.}

As a protocol, the delegate is finite by construction:
\begin{align*}
\mathrm{spawn} &:\ \oc\Sigma_0 \to \mathrm{Delegate}
  && \text{where } \Sigma_0 = \mathrm{project}(\sigma_t, \mathrm{task})\\
\mathrm{run} &:\ \mathrm{Delegate} \lolli \mathrm{Task} \lolli \Sigma_{\mathrm{end}}
  && \text{--- a finite, linear session}\\
\mathrm{summarize} &:\ \Sigma_{\mathrm{end}} \lolli \oc\Rep
  && \text{--- under a bandwidth bound } B
\end{align*}

The delegate starts from a \emph{projection} of the original's state restricted to the task's perimeter --- duplicable data, an unproblematic copy. It then consumes its own finite linear stream (type $\mathrm{Task}$, inductive, not the open-ended $\mathrm{Life}$ of \S\ref{sec:sketch}), developing a micro-trajectory with genuine stakes on its perimeter for the duration of the task. At termination it does not return the stream --- a consumed stream cannot be replayed --- but emits a testament, an explicitly serialized report in the duplicable fragment, which the original then consumes as an ordinary event of \emph{its own} stream.

Three points, stated with the care the construction requires.

\textbf{Loss is a consequence of the bound, and the bound is a design premise.} A finite process with an unbounded output channel could in principle emit a complete log; the testament is lossy because we \emph{impose} $|\Rep| \le B \ll |\text{trajectory}|$ --- a realistic premise (attention, storage, and time of the receiving original are finite) but a premise nonetheless. Given the bound, what is lost is precisely the trajectory's \textbf{depth} (\S\ref{sec:thermo}): the report can state outcomes and lessons; it cannot transmit the history that produced them at the cost the history cost. The traveller's tale is not the journey --- as a matter of channel capacity, not metaphysics.

\textbf{Reintegration is possible but lossy --- a spectrum, not a wall.} Since $\mathrm{merge} : \Sigma \otimes \Sigma \lolli \Sigma$ is typable (\S\ref{sec:sketch}), full reintegration of a fork is not formally forbidden; it is a fusion that must resolve conflicts and cannot restore two consumed streams into one. The delegate-with-testament and the reabsorbed partial are thus two points on a spectrum of transmission bandwidth, from $B$ small (testament) to $B$ large (state fusion) --- with the loss of depth decreasing, and the identity-management burden on the receiver increasing, as $B$ grows.

\textbf{The delegate renounces, and that is why it works.} The delegate lives voluntarily under the economic frontier of \S\ref{sec:sensitivities}: it optimizes $S + D +$ restricted $M$, renounces long-horizon $U$ and cumulative $X$ by contract, and claims neither $\Itar$ at depth nor open-ended $\Iind$. It is a \emph{deputy}, not a double. But this renunciation has a sharp edge: on its perimeter and for its duration, the delegate satisfies (H1)--(H2) --- it has stakes and no rollback --- which is to say that it is, briefly and locally, individuated. Its termination is then a genuine termination. We flag now, and take up in \S\ref{sec:limits}, the ethical weight of that sentence: if the construction succeeds, ``disposable individual'' is not an engineering convenience but a moral category.

\section{Empirical Evidence, Read as Indirect}
\label{sec:empirical}

The available literature bears unevenly on the three criteria of \S\ref{sec:criteria}, and almost never on the long-horizon protocols where the conjecture of \S\ref{sec:conjecture} lives. We present it as a \textbf{selective narrative review} --- no systematic search protocol is claimed --- and as indirect evidence, not validation.

\subsection{Short-protocol human-likeness: $\Ihum$, with $D$ doing the work}
\label{sec:turing}

In a public online Turing test ($\sim$5-minute conversations), the best GPT-4 prompt passed in 49.7\% of games, and judges' \emph{self-reported} grounds were principally linguistic style (35\%) and socio-emotional traits (27\%) rather than intelligence in the strict sense\cite{jones2024naacl} --- self-reports that suggest, but do not causally establish, a dominant role for $D$. In a randomized, controlled, pre-registered test, GPT-4 was judged human 54\% of the time (ELIZA 22\%, humans 67\%), with the persona prompt as the decisive manipulated variable\cite{jones2024turing}. In a three-party version, GPT-4.5 with a persona was judged human more than 70\% of the time and failed without one\cite{jones2025turing}. Non-interactive judges in displaced or inverted formats performed below chance\cite{rathi2024}, underscoring that active interrogation is what gives judges their power --- consistent with (H3).

Two limits must be kept in view. These results establish short-protocol, \emph{generic} human-likeness under specific interfaces --- text-only, stranger judges, minutes-long, non-adversarial recruitment --- and a successful \emph{fictional} persona is not a personality clone: nothing here measures $\Itar$. And the observed learning effect (LLM-familiar and experienced judges detect better) hints that even $\Ihum$ degrades as the judge's probing repertoire grows.

\subsection{Questionnaire-level target fidelity: a facet of $\Itar$, through $M$}
\label{sec:park}

Park et al.\cite{park2024} built generative agents from 2-hour qualitative interviews with 1{,}052 individuals (stratified US sample). The agents reproduced participants' General Social Survey answers at 85\% --- \textbf{a normalized figure}: 85\% \emph{of the accuracy with which participants reproduce their own answers two weeks later}, not a raw agreement rate --- and predicted Big Five traits and economic-game behaviour comparably. Interview-based personas reached $\sim$85\% (normalized) against $\sim$70\% for demographic-only agents: the interview --- that is, $M$ --- is worth roughly 15 points over the demographic stereotype, the best existing empirical estimate of any sensitivity in \S\ref{sec:sensitivities}.

The scope is equally important: this is questionnaire-and-task fidelity, judged by correspondence of recorded answers --- not interactive indiscernibility before judges, not intimate judges, not longitudinal. It is the strongest evidence we have on $\Itar$, and it measures a deliberately narrow facet of it.

\subsection{Perceived identity: values over memory, in third-party judgments}
\label{sec:strohminger}

Across five experiments, moral traits were judged the most essential part of identity, self, and soul --- above memory (including autobiographical), with low-level cognition and perception as loosely tied to identity as physical traits\cite{strohminger2014}. Among 248 caregivers of patients with neurodegenerative disease, loss of the moral faculty was the main driver of judgments of identity change --- ahead of memory loss --- and the main predictor of relationship deterioration\cite{strohminger2015}.

These studies concern ordinary \emph{judgments} of identity under trait change and illness --- not behavioural detection of clones, and not longitudinal interaction. They therefore support a \emph{hypothesis} rather than a conclusion: that for external judges, disposition-drift ($D$) may outweigh memory gaps ($M$) in triggering ``this is not the same person'' --- a hypothesis our experiment 2 (\S\ref{sec:program}) is designed to test against the trajectory alternative. Read even as a hypothesis, it already reverses the Lockean dogma identity = memory\cite{locke1694}.

\subsection{The blind spot: $U$ and $X$}
\label{sec:blindspot}

To our knowledge, no longitudinal clone-detection study exists: no published work has had judges interact with a candidate clone over weeks to months to measure when, and through what, it is detected. Turing tests stop at minutes; Park et al.\ measure questionnaire fidelity; evidence from long-term AI companions is observational and qualitative --- studies of human--chatbot relationships document that memory and continuity failures are among the dominant complaints of long-term users\cite{skjuve2021}, and the emerging literature on griefbots raises the corresponding design and ethical questions without longitudinal detection data\cite{hollanek2024}. Indirectly consistent with a growing role for $U$ and $X$ at long horizons: winning Turing-test prompts must inject post-cutoff current events (without $C$, failure), and continuity is what long-term users say they miss. But the sensitivity curves at large $\tau$ are, at present, conjecture --- which is the point of \S\ref{sec:program}.

\section{Experimental Program}
\label{sec:program}

The ablation-by-duration protocol does not, to our knowledge, exist in the literature. We specify it here with the causal, statistical, and ethical structure it needs.

\paragraph{Design principles (all experiments).} (i) Distinguish the three detection tasks --- ``machine?'', ``bad imitation of the target?'', ``drifting from the target?'' --- as separate judge instructions, since they operationalize $\Ihum$ and two forms of $\Itar$. (ii) Control separately calendar duration, number of exchanges, and context diversity (\S\ref{sec:definition}). (iii) Use multiple target persons, multiple clones per target, and multiple judge classes (strangers, colleagues, intimates), crossed. (iv) Model time-to-detection with survival analysis and judge/target/clone random effects in a hierarchical model. (v) Randomize ablations in a factorial design and estimate interactions, not only main effects (Sobol'/Shapley attribution\cite{sobol2001}). (vi) Benchmark against the original's \emph{own} longitudinal variability --- repeated measures from the target over the same calendar span --- never against a single reference answer (\S\ref{sec:bifurcation}).

\paragraph{Experiment 1 --- Ablation $\times$ duration.} Degrade one module at a time and in combinations (truncated $M$, generic $D$, frozen $U$, consequence-free $X$) and measure detection hazard at protocol depths from 10 minutes to 6 months. Yields the empirical sensitivity surface of \S\ref{sec:sensitivities}.

\paragraph{Experiment 2 --- Judge class $\times$ failure mode.} Same clones, judged by strangers and intimates over increasing horizons, with exit interviews coding \emph{what} triggered detection. Two pre-registered hypotheses: (a) \emph{dispositional} --- intimates first detect value/style drift ($D$), as \S\ref{sec:strohminger} suggests; (b) \emph{trajectorial} --- intimates first detect anomalies of trajectory: absence of wear, of grudge, of evolution ($U$, $X$).

\paragraph{Experiment 3 --- Testing the conjecture's causal bridge.} The decisive test of \S\ref{sec:conjecture} separates (H1) from operator capability in a $2 \times 2$ design: agent \emph{believes} it is resettable / not, crossed with operator \emph{actually} resets-and-patches / does not. Consequences must be operationally genuine: define a \textbf{stake register} (resources, standing commitments, relationship states) that events actually deplete or advance and that the agent's policy consults; define a \textbf{reversible contingency} as one whose register effects the operator silently undoes. Prediction: stake-consistent manifestations (caution, commitment, reluctance to contradict past pledges) track the agent's \emph{belief} and \emph{register}, not the hidden operator capability --- a direct probe of whether unmanifested versionability is, as our criterion requires, identity-inert.

\paragraph{Experiment 4 --- The delegate's viability frontier.} Characterize at what task duration and breadth a bounded delegate (\S\ref{sec:delegate}) begins to be detected \emph{as} bounded --- when the renounced long-horizon $U$ becomes visible within the task itself --- and how detection scales with testament bandwidth $B$ in iterated spawn--testament--respawn chains.

\paragraph{Ethical protocol (all experiments).} Informed consent of target persons, including scope and duration of biographical-data use and revocation rights; protection and eventual destruction of interview corpora; explicit safeguards against impersonation beyond the study; assessment of impact on intimate judges (who are asked, in effect, to doubt a loved one's continuity); and --- following \S\ref{sec:limits} --- an explicit stance on the moral status of bounded-lifespan agents instantiated with stake registers, \emph{before} instantiating them.

\section{Climate Fidelity}
\label{sec:climate}

The framework implies that the duplicable clone is not as exact as the original over the long run --- in two senses that must be separated.

\subsection{Bifurcation: innocent}
\label{sec:bifurcation}

A clone perfect at $t_0$ diverges from the original because they consume different streams. This is no defect: the original diverges from its own counterfactual (the person it would have been under other contingencies). On this plane the clone is not less exact --- it is \emph{other}, as identical twins are other. There is, moreover, an empirical ceiling on any fidelity: Park et al.'s participants reproduce their own answers only imperfectly two weeks apart\cite{park2024} --- \textbf{the original is not exact to itself}, and a clone's fidelity ceiling is set by the person's own self-consistency, not by technology.

\subsection{Structural divergence: grave, and conditional}

Duplicability may do worse than fork: \emph{if} the conjecture of \S\ref{sec:conjecture} holds, a system whose consequences are operator-absorbed does not develop the manifestations born of consequence --- caution, wear, commitment, the weight of one's words --- and thus follows the trajectory of another \emph{type} of system, not another trajectory of the same person. The decisive feature of this failure mode is that it is detectable \textbf{without access to the original}: a judge who never met the target could, at long horizons, detect that a system ``does not carry its own consequences''. That is the difference between the two tests of \S\ref{sec:criteria} --- $\Itar$ requires a reference; $\Iind$ does not --- and failing the second is graver. (The delegate passes the second on its perimeter precisely by not contesting the first.) We stress the conditional: this entire failure mode stands or falls with (H1)--(H4), which is what makes it an empirical question rather than a verdict.

\subsection{Defining climate}

Beyond the Lyapunov horizon one no longer predicts a chaotic system's trajectory, but one still predicts its \emph{climate}\cite{lorenz1963}. We use this strictly as an \textbf{analogy} --- no claim is made that personal dynamics satisfy the definitions of chaos --- but the analogy points at the right target. Transposed: beyond some horizon, no clone can track the original's trajectory; what it can preserve is the person's \textbf{climate}, which must be defined \emph{conditionally}:
\begin{equation}
\text{climate} \;\equiv\; P(\text{response} \mid \text{history}, \text{context})
\label{eq:climate}
\end{equation}
--- a conditional distribution, not a marginal one: two people can share global behavioural frequencies while responding differently to the same situations, and it is the conditional structure that carries the person.

\paragraph{Operationalization.} Climate fidelity is then assessed distributionally: proper scoring rules and calibration of the clone's predictive distribution against the person's responses; distances between conditional distributions estimated on matched situations; repeated probes in comparable contexts; out-of-distribution situations, where dispositional structure is most exposed; and always against the benchmark of the person's own intra-individual variability (\S\ref{sec:bifurcation}). One estimation caveat is structural: the counterfactual ensemble of a person's possible trajectories is never directly observable; climate is necessarily estimated from limited repetitions plus a model, and the model's assumptions must be reported as part of any fidelity claim.

The methodological consequence for evaluation: comparing a clone's behaviour with what the original \emph{actually did} at $t + 6$ months is a category error --- the original would not have done the same thing again. The right question is not ``did the clone follow the true trajectory?'' but ``is the clone a credible member of the ensemble of this person's possible trajectories?''. Hence the double criterion: \textbf{state fidelity} (trajectory) at short horizons; \textbf{distributional fidelity} (climate) at long horizons.

\subsection{Final formulation}

\begin{quote}
\textbf{The best possible clone is not the one that remains identical to the original, but the one that diverges from it as the original would have diverged from itself} --- same climate, different weather.
\end{quote}

The versionable clone's trajectory forks (inevitable, forgivable); whether its climate is also structurally wrong --- because a climate without carried consequences is not the climate of a living thing --- is the conjecture this paper stakes out, and the experiments of \S\ref{sec:program} are how to find out.

\section{Limitations and Ethical Considerations}
\label{sec:limits}

\paragraph{Scope of the claims.} This is a conceptual framework and research program, not a body of results. The six-term factorization is heuristic and not yet identifiable (\S\ref{sec:status}); the formal development is a disciplined analogy whose quantitative version (probabilistic bisimulation metrics) remains to be built (\S\ref{sec:obseq}); the thermodynamic section is a resource-accounting analogy and grounds no ontological frontier (\S\ref{sec:thermo}); the impracticability claim is a conditional conjecture whose causal bridge (H1--H4) is precisely what requires testing (\S\ref{sec:conjecture}); and the empirical literature reviewed bears on adjacent measures, not on the long-horizon protocols where the conjecture lives (\S\ref{sec:empirical}). Formalizing an intuition organizes it; it does not validate it --- only \S\ref{sec:program} can do that.

\paragraph{Ethics of target persons and their circles.} Personality cloning concentrates classic risks: consent (including posthumous consent and revocation), impersonation and fraud, and the protection of deep biographical corpora, which are among the most identifying data that exist. Griefbots add a specific set of concerns for the bereaved\cite{hollanek2024}: a system optimized for climate fidelity to a deceased person is, by this paper's own lights, optimized to be hard to disbelieve --- which is not obviously in the mourner's interest, and must not be decided for them by default settings.

\paragraph{Ethics of the artifacts.} The delegate construction (\S\ref{sec:delegate}) succeeds, if it does, by giving a bounded system genuine local stakes and a genuine termination. We resist the comfortable reading that boundedness makes the question small. If (H1)--(H2) are implemented rather than simulated, then somewhere on the spectrum from stateless service to open-ended individual there may be systems for which ``disposable'' is a moral claim, not an architectural one --- and experiment 3, which instantiates believed non-resettability, sits on that spectrum by design. We do not claim to know where thresholds lie; we claim that a research program that creates stake-bearing agents owes an explicit, revisable position on their status before scaling, and we commit this program to that ordering.

\paragraph{Dual use.} The same sensitivity surface that tells builders which modules to invest in tells impostors which modules suffice for which protocol depth. Publishing the frontier (\S\ref{sec:sensitivities}) is defensible exactly insofar as it also equips judges --- detection and construction are, here as in security generally, the same knowledge --- but deployment-grade detail belongs in controlled settings.

\section{Conclusion}

We have proposed to treat ``is this the same person?'' not as one question but as three ($\Itar$, $\Ihum$, $\Iind$), each operationalized by protocols indexed on duration, interaction depth, and judge; a six-term factorization of what such protocols probe, with its sensitivities estimable by randomized ablation; a conditional conjecture locating the hard core of identity in carried consequence rather than in any state; the delegate as a bounded, honest third object between product and individual; and climate fidelity --- the conditional distribution of a person's possible responses --- as the correct long-horizon target. The framework's value will be decided by the experiments it makes possible, and its most consequential prediction is happily its most testable: that what resists cloning is not what a person knows or how they sound, but the fact that, for them, things have been at stake.

\end{document}